%% file: main.tex
\documentclass[10pt,twocolumn,letterpaper]{article}

\usepackage[pagenumbers]{cvpr} 

\input{preamble}

\definecolor{cvprblue}{rgb}{0.21,0.49,0.74}
\usepackage[pagebackref,breaklinks,colorlinks,allcolors=cvprblue]{hyperref}
\usepackage{booktabs}
\usepackage{multirow}
\def\paperID{*****} 
\def\confName{CVPR}
\def\confYear{2026}

\title{MUT3R: Motion-aware Updating Transformer for 
Dynamic 3D Reconstruction}

\author{
Guole Shen\textsuperscript{1}*,
Tianchen Deng\textsuperscript{1}*,
Xingrui Qin\textsuperscript{1},
Nailin Wang\textsuperscript{1},
Jianyu Wang\textsuperscript{2},
Yanbo Wang\textsuperscript{1},\\
Yongtao Chen\textsuperscript{1},
Hesheng Wang\textsuperscript{1},
Jingchuan Wang\textsuperscript{1} \\
\textsuperscript{1} Shanghai Jiao Tong University 
\textsuperscript{2} Nanyang Technological University
}

\begin{document}
\maketitle
\renewcommand{\thefootnote}{} 
\footnotetext{ The first two authors contribute equally to this paper. Corresponding author: Jingchuan Wang (jchwang@sjtu.edu.cn)}

\input{sec/0_abstract}    
\input{sec/1_intro}

\input{sec/2_related_work}
\input{sec/3_method}

\input{sec/4_experiment}

\input{sec/5_ablation_study}
\input{sec/6_conclusion}

\clearpage

{
    \small
    \bibliographystyle{ieeenat_fullname}
    \bibliography{main}
}


\end{document}

%% file: preamble.tex
\usepackage[table]{xcolor}



%% file: sec/0_abstract.tex
\begin{abstract}

Recent stateful recurrent neural networks have achieved remarkable progress on static 3D reconstruction but remain vulnerable to motion-induced artifacts, where non-rigid regions corrupt attention propagation between the spatial memory and image feature.
By analyzing the internal behaviors of state and image token updating mechanism, we find that aggregating self-attention maps across layers reveals a consistent pattern: dynamic regions are naturally down-weighted, exposing an implicit motion cue that the pretrained transformer already encodes but never explicitly uses.
Motivated by this observation, we introduce MUT3R, a training-free framework that applies the attention-derived motion cue to suppress dynamic content in the early layers of the transformer during inference. Our attention-level gating module suppresses the influence of dynamic regions before their artifacts propagate through the feature hierarchy.
Notably, we do not retrain or fine-tune the model; we let the pretrained transformer diagnose its own motion cues and correct itself.
This early regulation stabilizes geometric reasoning in streaming scenarios and leads to improvements in temporal consistency and camera pose robustness across multiple dynamic benchmarks, offering a simple and training-free pathway toward motion-aware streaming reconstruction.
\vspace{-0.1cm}
\end{abstract}

%% file: sec/1_intro.tex
\section{Introduction}
\label{sec:intro}

\begin{figure}[t]
    \centering
    \includegraphics[width=\linewidth]{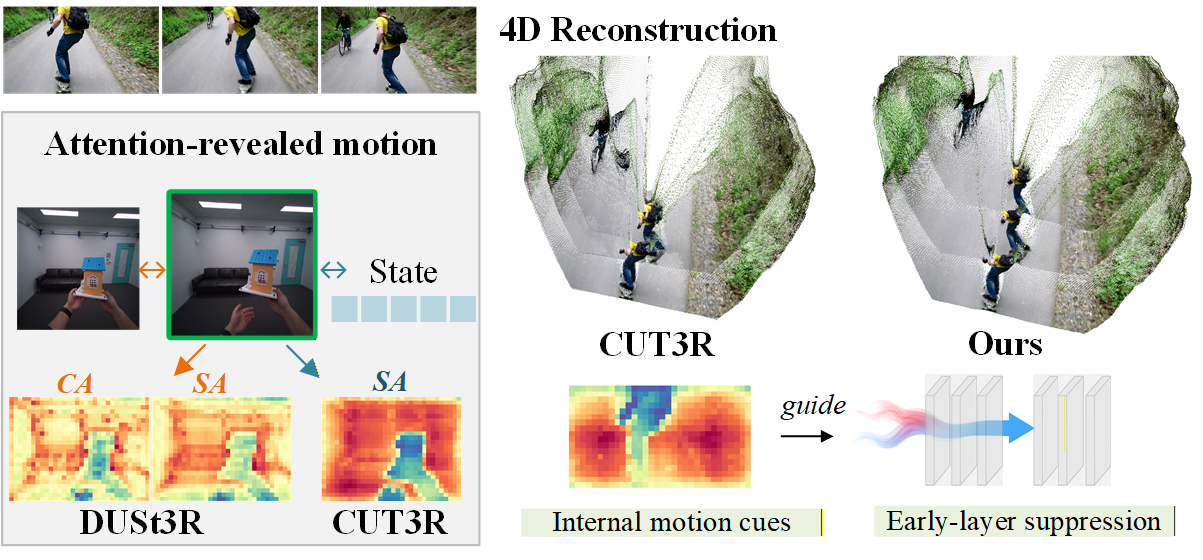}
    \caption{
We present MUT3R, a training-free framework that derives motion cues from CUT3R’s attention maps and performs early-layer motion suppression to enhance robustness in dynamic scenes; unlike pairwise DUSt3R, which requires extra temporal fusion, the streaming design of CUT3R carries historical context in its state tokens, making such dynamics directly reflected in self-attention. (SA denotes self-attention and CA denotes cross-attention in the figure.)
    }
    \label{fighead}
    \vspace{-0.5cm}
\end{figure}

Reconstructing 3D geometry from images underpins a wide range of downstream applications in novel view synthesis, AR/VR, autonomous navigation, and robotics. Classical Structure-from-Motion~\cite{schonberger2016structure,agarwal2011building} and Multi-view Stereo pipelines~\cite{huang2018deepmvs,yao2018mvsnet} address this problem by decomposing it into correspondence matching, pose estimation, and depth fusion. While effective in static and well-conditioned environments, these optimization-heavy methods struggle with textureless regions, illumination changes, and dynamic content, motivating the shift toward more unified, feed-forward transformer models.
Recently, transformer-based feed-forward reconstruction models have demonstrated impressive capability in estimating camera poses and dense geometry directly from images.

Methods such as DUSt3R~\cite{wang2024dust3r} and MASt3R~\cite{leroy2024grounding} recover dense 3D-consistent structure by regressing pixel-aligned pointmaps from image pairs, establishing reliable cross-view correspondences without explicit geometric supervision.
However, these models operate in a pairwise setting, where only two frames are jointly processed and rely on a global alignment method to align the predicted pointmap into the global map. Some works like CUT3R~\cite{wang2025continuous} and Spann3R~\cite{wang20253d} further improve the framework to address streaming inputs with a continuous update state and an external memory database.
Although CUT3R is trained on dynamic sequences, dynamic content can still inject disturbances into the spatial memory, degrading static-region reconstruction and pose stability. This issue arises from the inherent structural behavior of feed-forward 3D transformers, rather than from limitations in the data or supervision.
To understand this behavior, we revisit CUT3R’s internal mechanism.
As a representative feed-forward transformer for streaming 3D reconstruction, its decoder continually exchanges information between image tokens and state tokens, which act as a latent temporal memory.
Through this process, we observe that temporally stable regions tend to reinforce consistent correspondences, whereas dynamic regions struggle to maintain alignment, producing characteristic fluctuations in self-attention.
This reveals that \textbf{the pretrained decoder naturally exhibits motion-sensitive attention patterns}, even without any explicit motion module.
Such patterns constitute an implicit motion signal that has not yet been explicitly exploited. 
Motivated by this insight, we introduce MUT3R, a training-free attention adaptation framework that enhances CUT3R’s sensitivity to dynamics without modifying its architecture or retraining its weights.
MUT3R first analyzes multi-layer self-attention responses from the frozen decoder and aggregates them into a patch-wise dynamic score map that captures motion-driven fluctuations in attention.
This map is then injected as a soft bias into the early decoder layers via attention-level gating mechanism. 
The gating selectively attenuates unstable queries or keys depending on the attention direction---image self-attention, state-to-image cross-attention, or image-to-state cross-attention---thereby preventing motion interference from propagating forward. 
Because later layers already exhibit strong geometric consistency, early-layer suppression becomes a structurally targeted intervention that preserves CUT3R’s pretrained 3D reasoning while filtering out transient motion effects.
Across dynamic reconstruction benchmarks, MUT3R consistently improves depth stability, pose robustness, and long-range temporal coherence. \textbf{Overall, our contributions are shown as follows:}
\begin{itemize}
    \item \textbf{Unsupervised motion cues discovery.}
    We observe that pretrained feed-forward 3D transformers implicitly encode motion saliency within their attention maps, providing a reliable internal motion probe without requiring any external supervision.

    \item \textbf{Training-free dynamic information suppression.}
    We introduce a attention gating mechanism that uses dynamic cues to suppress motion interference in early decoder layers, without adding additional network structure or fintuning the network.

    \item \textbf{Concise and Effective.}
   Our simple attention enhancement consistently improves robustness in dynamic environments, yielding more stable video depth, more accurate camera poses, and more temporally consistent 4D reconstruction across diverse benchmarks.
\end{itemize}

%% file: sec/2_related_work.tex
\section{Related Work}

\begin{figure*}[t]
    \centering
    \includegraphics[width=\linewidth]{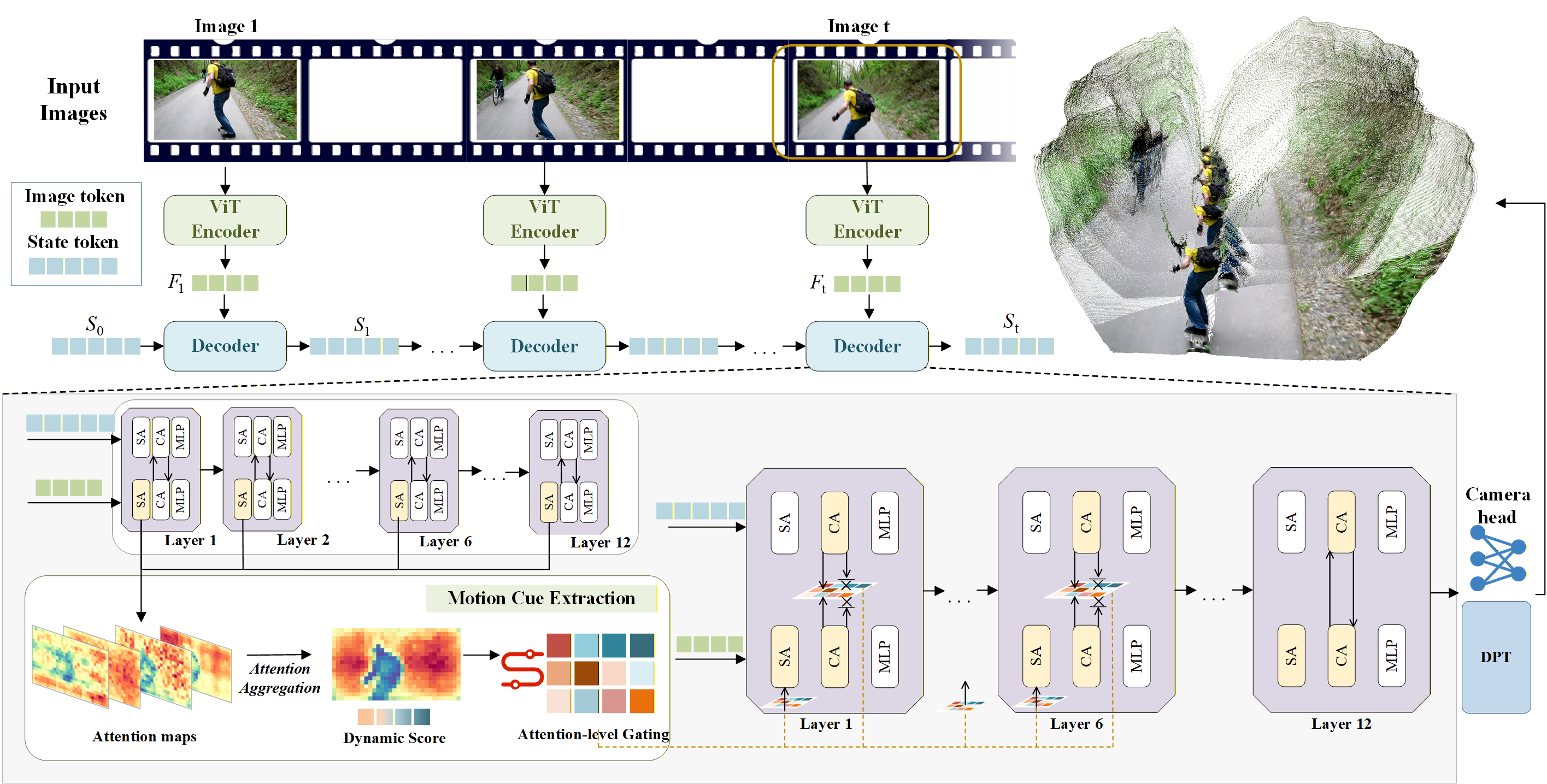}
    \caption{
\textbf{Method pipeline.}
Given a sequence of input frames, each image is first encoded into visual tokens $F_t$ by a frozen ViT encoder.
The tokens are processed by the recurrent CUT3R decoder with a persistent state.
Inside each decoder layer, self-attention (\textbf{SA}) and cross-attention (\textbf{CA}) jointly refine the image and state tokens.
We analyze the attention responses from multiple layers to estimate a patch-level dynamic score map.
These scores guide our \textbf{attention-level gating} module that suppresses dynamic regions during early-layer attention.
The resulting motion-aware updating enables robust and consistence 3D reconstruction without finetuning the base model.
}
\vspace{-0.4cm}
\label{pipeline}
\end{figure*}

\paragraph{Classical SfM and SLAM.}
Classical 3D reconstruction is primarily accomplished through Structure-from-Motion (SfM) from unordered images and visual SLAM from video sequences, both estimating camera trajectories and sparse or semi-dense 3D structure by enforcing multi-view geometric consistency~\cite{schonberger2016structure,campos2021orb}. Feature-based pipelines detect and describe local features, match them across images, filter correspondences via robust estimation and minimal solvers, and refine cameras and structure with bundle adjustment~\cite{agarwal2010bundle,agarwal2011building,detone2018superpoint,sarlin2020superglue,Deng_2024_CVPR,shen2025grs}. Direct and semi-direct methods instead minimize photometric error over poses and dense geometry, with recent deep SLAM and visual odometry (VO) systems integrating learned feature encoders with differentiable bundle adjustment for accurate real-time trajectory estimation~\cite{engel2014lsd,engel2017direct,teed2021droid,hidalgo2022event,teed2023deep,wang2024vggsfm,deng2025mne,deng2025mcnslammultiagentcollaborativeneural,neslam}. Despite their maturity, these pipelines depend on multi-stage optimization and strong static-scene assumptions, making them fragile under motion, occlusion, and illumination changes.
\vspace{-0.5cm}
\paragraph{Feed-Forward 3D Reconstruction under Static Assumptions.}
The introduction of DUSt3R marked a turning point for feed-forward 3D reconstruction.
By treating multi-view geometry as the regression of 2D-aligned pointmaps, it enabled direct estimation of camera poses and 3D structure in a single forward pass~\cite{wang2024dust3r}, significantly simplifying the traditional correspondence-and-refinement pipeline. MASt3R~\cite{leroy2024grounding} extends DUSt3R by jointly learning dense matching and 3D reconstruction, yielding stronger multi-view consistency and more reliable feed-forward geometry estimation. VGGT~\cite{wang2025vggt} provides a unified visual–geometry backbone trained jointly on pose, depth, and tracking signals, offering a foundation model that transfers effectively to a wide range of geometric tasks. While highly capable on static multi-view inputs, these feed-forward models lack mechanisms to handle non-rigid motion, occlusion changes, and other dynamic effects, limiting their applicability in dynamic scene reconstruction.
\vspace{-0.5cm}
\paragraph{Reconstructing Dynamic Scenes from Monocular Videos.}
Reconstructing dynamic scenes is challenging due to non-rigid motion, occlusion, and motion blur. Existing approaches either jointly optimize geometry and motion from video~\cite{kopf2021robust,zhang2022structure,li2025megasam,lu2025align3r}, or learn neural 4D representations that model dynamic radiance fields or scene flow~\cite{pumarola2021d,park2021hypernerf,li2021neural,gao2022monocular}.  Recently, several works have adapted foundation models to dynamic scenes. MonST3R fine-tunes the DUSt3R architecture on dynamic videos~\cite{zhang2025monst3r}, and DAS3R introduces learned dynamic masks with dedicated reconstruction branches~\cite{xu2024das3r}.
Other work avoids retraining: Easi3R~\cite{chen2025easi3r} performs training-free attention adjustment within the pairwise DUSt3R framework, whose attention reflects only two-frame differences and therefore requires additional temporal fusion.
In contrast, CUT3R’s state tokens retain temporal history, producing stable motion cues in self-attention that we directly leverage without extra temporal fusion.

\paragraph{Continuous Reconstruction Methods.}
In long-horizon and interactive scenarios, frames arrive sequentially and the system must update geometry online. CUT3R and TTT3r maintains a compact recurrent state that integrates new observations into a shared coordinate system\cite{wang2025continuous,ttt3r}
. Spann3R uses an external spatial memory to predict per-image pointmaps directly in a global frame, and Point3R employs a spatial pointer memory anchored at 3D positions for better retention of distant observations~\cite{wang20253d,wu2025point3r}. STREAM3R formulates pointmap prediction with causal attention, key--value caching, and windowed contexts so that sequential inference scales to long videos with controlled compute~\cite{lan2025stream3r}. MUT3R is complementary to these streaming architectures: it keeps network weights fixed and introduces a lightweight, training-free motion-aware attention bias that can be plugged into existing streaming pointmap transformers.

%% file: sec/3_method.tex
\section{Method}


Recent transformer-based feed-forward frameworks for 3D reconstruction~\cite{wang2024dust3r} rely on alternating self-attention and cross-attention layers that iteratively align and refine image features. These methods achieve high accuracy 3D reconstruction in static scenes, while have difficulties in dynamic scenes. To address this limitation, Easi3R~\cite{chen2025easi3r} introduces a pair-wise attention adaptation module designed for dynamic objects. While multi-view inputs and spatial memory naturally provide stronger cues for dynamic perception, this regime is still underexplored and current methods often fail to maintain consistency in long-sequence scene reconstruction. We therefore build upon CUT3R as our backbone and systematically examine how dynamic content affects multi-view streaming reconstruction.
The state tokens in CUT3R are serve as latent temporal memory, guiding image tokens through cross-attention.
Although the state tokens do not correspond to explicit physical entities, they act as latent temporal memories whose interaction with image tokens allows the model to associate geometrically consistent regions across time, implicitly capturing motion and stability cues.
As a result, we observe that the image self-attention maps begin to exhibit motion-sensitive behavior, where dynamic regions often lead to unstable or dispersed attention, while static regions maintain concentrated and consistent responses.

This phenomenon motivates our method.
As shown in Figure~\ref{pipeline}, we directly analyze the self-attention maps of the pretrained CUT3R decoder to infer patch-level motion cues.
These cues reveal which image regions behave consistently across frames and which fluctuate due to motion.
Building on this insight, MUT3R introduces a training-free, plug-and-play early-layer motion suppression mechanism that derives motion cues from self-attention and uses them to attenuate motion-dominant responses.
By filtering such unstable activations at early layers, MUT3R improves robustness and temporal stability in dynamic scenes—without any retraining or architectural modification.

\begin{figure}[t]
    \centering
    \includegraphics[width=\linewidth]{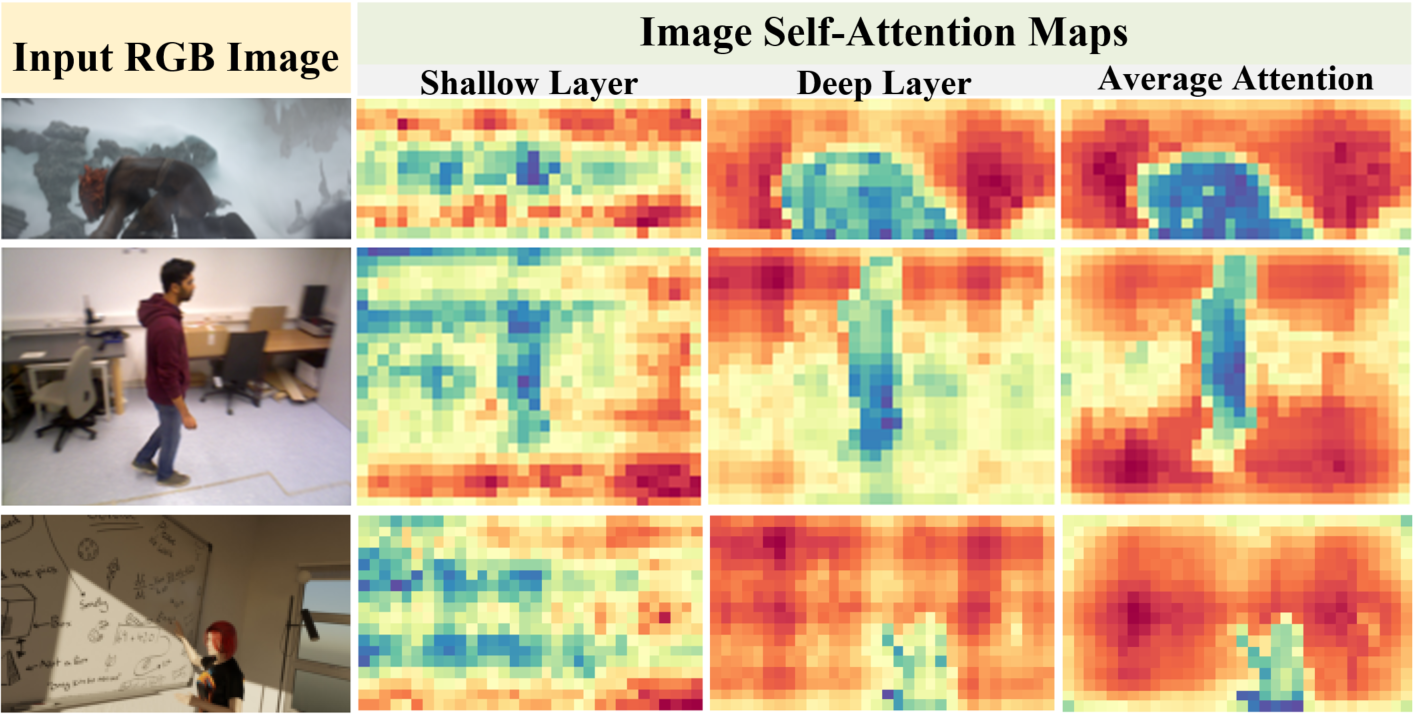}
    \caption{
        \textbf{Visualization of different level attention maps.}
       From left to right, we show the input RGB frame, self-attention responses from shallow layer and deep decoder layer, and the averaged attention map across all layers. Blue indicates lower attention responses, while red indicates higher responses.
    }
    \vspace{-0.2cm}
    \label{dynlatent}
\end{figure}

\subsection{Architecture Overview: CUT3R Backbone}

Our framework builds directly upon the frozen CUT3R~\cite{wang2025continuous} architecture, 
which reconstructs 3D geometry from sequential video frames through a recurrent transformer decoder. 
At each time step $t$, the system receives an input image $\mathbf{I}_t$, 
which is first encoded into token representations $\mathbf{F}_t$ through a vision transformer encoder~\cite{yuan2021tokens}:
\begin{equation}
\mathbf{F}_t = \text{Encoder}(\mathbf{I}_t),
\end{equation}
where $\mathbf{F}_t \in \mathbb{R}^{N\times C}$ denotes the patch-level embeddings.

To integrate the updated memory with current observations, 
we adopt a transformer-based decoder in which the memory tokens interact 
with the current frame features $\mathbf{F}_t$ and the pose token $\mathbf{z}_t$:
\begin{equation}
[\mathbf{S}_t,\, \mathbf{z}'_t \oplus \mathbf{F}'_t]
= 
\text{Decoder}\!\left([\mathbf{S}_{t-1},\, \mathbf{z}_t \oplus \mathbf{F}_t]\right),
\end{equation}
where $\mathbf{S}_t$ denotes the updated memory state, 
$\mathbf{F}'_t$ are the enriched frame features, 
and $\mathbf{z}'_t$ is a dedicated pose token capturing global scene-level information. 
The decoder alternates between four complementary attention operations. 
Image and state tokens first undergo their own self-attention, 
followed by bidirectional cross-attention: 
image-to-state attention injects current visual evidence into the memory, 
while state-to-image attention propagates global geometric context back to the frame. 
These interactions produce the updated state tokens $\mathbf{S}_t$ 
and the enriched image tokens $\mathbf{F}'_t$, 
where $\mathbf{S}_t$ is carried forward as persistent memory 
and $\mathbf{F}'_t$ supports depth, pose, and 3D reconstruction.  

Based on $\mathbf{F}'_t$ and $z'_t$, CUT3R predicts metric-scale 3D pointmaps $X$ in both the camera and world frames, as well as the 6-DoF camera pose $P$:
\begin{equation}
\hat{X}^{\text{self}}_t, \hat{C}^{\text{self}}_t 
= \text{Head}_{\text{self}}(\mathbf{F}'_t),
\end{equation}
\begin{equation}
\hat{X}^{\text{world}}_t, \hat{C}^{\text{world}}_t 
= \text{Head}_{\text{world}}(\mathbf{F}'_t, z'_t),
\end{equation}
\begin{equation}
\hat{P}_t = \text{Head}_{\text{pose}}(z'_t),
\end{equation}
where $\text{Head}_{\text{self}}(\cdot)$ and $\text{Head}_{\text{world}}(\cdot)$ 
use DPT~\cite{dpt}, 
and $\text{Head}_{\text{pose}}(\cdot)$ is an MLP.  
All outputs are in metric scale.

Importantly, our method does not modify the CUT3R architecture or retrain it from scratch.
Instead, we analyze the attention maps inside the decoder 
and introduce a training-free gating mechanism that adaptively suppresses motion-dominated regions.

\subsection{Self-Attention Driven Motion Cue Extraction}

We analyze the internal self-attention maps of the frozen CUT3R~\cite{wang2025continuous} decoder to reveal how motion cues emerge without explicit supervision. 
Let the image feature tokens be $\mathbf{F}\!\in\!\mathbb{R}^{N\times C}$, 
where $N$ is the number of tokens, and $C$ the feature dimension. 
The query, key, and value projections are obtained by linear layers:
\begin{equation}
\mathbf{Q}=\ell_Q(\mathbf{F}),\quad 
\mathbf{K}=\ell_K(\mathbf{F}),\quad 
\mathbf{V}=\ell_V(\mathbf{F}),
\end{equation}
with $\mathbf{Q},\mathbf{K},\mathbf{V}\in\mathbb{R}^{H\times N\times C_h}$, 
where $H$ denotes the number of attention heads and $C_h=C/H$. 
The attention map at layer $l$ is computed as:
\begin{equation}
\mathbf{A}_l=\text{Softmax}\!\left(\frac{\mathbf{Q}_l\mathbf{K}_l^{\top}}{\sqrt{C_h}}\right),
\quad \mathbf{A}_l\!\in\!\mathbb{R}^{H\times N\times N}.
\end{equation}
To obtain a global motion prior, we average the attention maps 
along the batch, head, and key dimensions:
\begin{equation}
\bar{\mathbf{A}}
= \frac{1}{H L N_k}
\sum_{l=1}^{L}\sum_{h=1}^{H}\sum_{k=1}^{N_k}
\mathbf{A}_{l,h,:,k},
\end{equation}
resulting in a single $N_q$-dimensional vector that represents 
the average attention strength of each query token across all layers and heads.
Averaging over the key dimension captures how broadly each query attends to the rest of the image:
tokens with dispersed or unstable attention correspond to dynamic regions,
while those with concentrated attention indicate stable, static areas.
This aggregated vector thus serves as a coarse \emph{motionness prior}.

As illustrated in Fig.~\ref{dynlatent}, the self-attention responses exhibit a clear relationship with scene dynamics: shallow layers tend to react to local appearance and texture changes, while deeper layers-—after repeated interaction with the persistent state—produce more temporally stable patterns aligned with the underlying 3D structure.
By aggregating these multi-layer attention responses, we obtain a coarse motionness prior that captures this implicit motion awareness encoded by CUT3R, even without any explicit dynamic supervision.

To transform this aggregated response into a usable dynamic score, we apply a sigmoid normalization $g=\sigma(\bar{A})$.  
The sigmoid both rescales $\bar{A}$ into a stable $[0,1]$ range and sharpens the contrast between coherent, static tokens and unstable, motion-affected tokens—pushing them toward 0 and 1 respectively. This yields a more discriminative yet still soft dynamic score suitable for attention gating.

\subsection{Attention-Based Motion Suppression}

\begin{figure}[t]
    \centering
    \includegraphics[width=\linewidth]{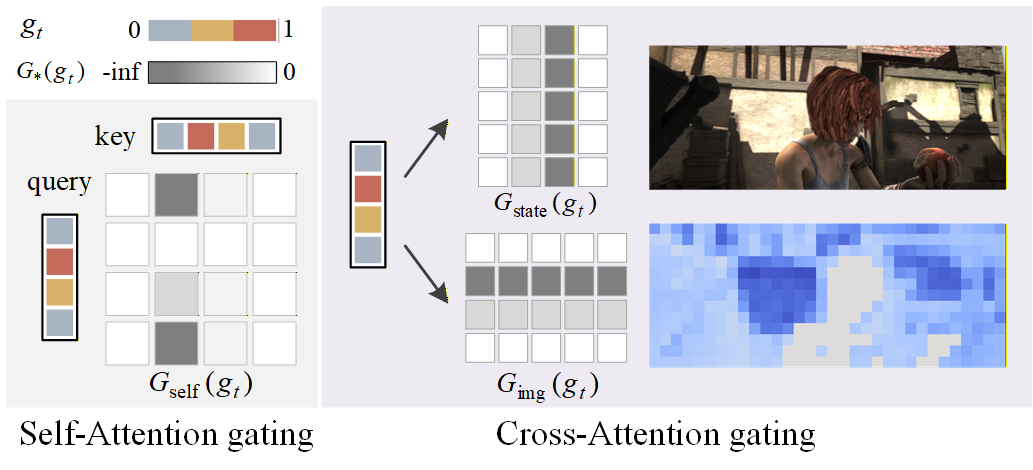}
\caption{
\textbf{Motion awareness attention gating.}
This figure illustrates how the dynamic score map $g_t$ is converted into attention biases that suppress motion-dominated regions. 
The left part shows self-attention gating. 
The right part visualizes the bias terms for both directions of cross-attention between image and state tokens. 
The picture depicts the gated update applied to the image tokens, highlighting where our motion-aware suppression reduces state-induced changes in dynamic regions.
}
\vspace{-0.5cm}
    \label{attn}
\end{figure}

To make the attention motion-aware, we define a unified gating function 
$\mathcal{G}_{\ast}(g_t)$ that adjusts the attention logits based on the dynamic score $g_t$. Figure~\ref{attn} provides an overview of this mechanism, illustrating how $g_t$ is converted 
into additive attention biases for self-attention and the two directions of cross-attention. The gated attention is formulated as:
\begin{equation}
\tilde{\mathbf{A}}_l = 
\text{Softmax}\!\left(
\frac{\mathbf{Q}_l \mathbf{K}_l^{\top}}{\sqrt{d}} 
+ \mathcal{G}_{\ast}(g_t)
\right),
\end{equation}
where $\ast \in \{\text{self},\, \text{state},\, \text{img}\}$ denotes the attention type.
For the \textbf{self-attention}, the gating term suppresses static queries from attending 
to dynamic keys:
\begin{equation}
\mathcal{G}_{\text{self}}(g_t)
= \beta\, \log\!\big(1 - (1 - g_t^{(q)})\, g_t^{(k)}\big).
\end{equation}

\noindent where $g_t^{(q)}$ and $g_t^{(k)}$ denote the motion scores of the query and key patches, 
both derived from the same dynamic score map $g_t$.  
This gating operates in a pairwise manner over all query–key pairs 
(\emph{i.e.}, an $N\times N$ modulation matrix), 
and thus dynamically suppresses static–to–dynamic interactions within the image tokens.

When updating the state $\mathbf{S}_t$, the gating function penalizes dynamic image regions:
\begin{equation}
\mathcal{G}_{\text{state}}(g_t) 
= \beta\, \log\!\big(1 - g_t^{(k)}\big),
\end{equation}
where dynamic \emph{keys} in the image are suppressed in a per-key fashion 
(\emph{i.e.}, a $1\times N_k$ modulation broadcast across queries).
In the reverse direction (image $\leftarrow$ state), the gating is defined as:
\begin{equation}
\mathcal{G}_{\text{img}}(g_t) 
= \beta\, \log\!\big(1 - g_t^{(q)}\big),
\end{equation}
which down-weights dynamic \emph{queries} in a per-query manner 
(\emph{i.e.}, an $N_q\times 1$ modulation broadcast along the key dimension).
This asymmetric design allows each attention direction to adaptively suppress 
motion-dominated tokens while preserving stable correspondences.

The hyperparameter $\beta$ controls the gating sharpness.
Specifically, intermediate values of $g_t$ lead to partial attenuation, 
allowing soft and differentiable control rather than hard masking.  
This unified soft-mask formulation effectively suppresses static–dynamic interference.  
The motionness map $g_t$ thus serves as a motion-aware routing weight that 
softly balances dynamic suppression.  
This asymmetric gating mirrors the directional nature of information flow: 
when transferring information from image to state, unstable image regions are suppressed; 
when injecting state information into the image, unstable queries are down-weighted.  
Together, they maintain temporal consistency without freezing dynamics entirely.

\begin{figure*}[t]
    \centering
    \includegraphics[width=\linewidth]{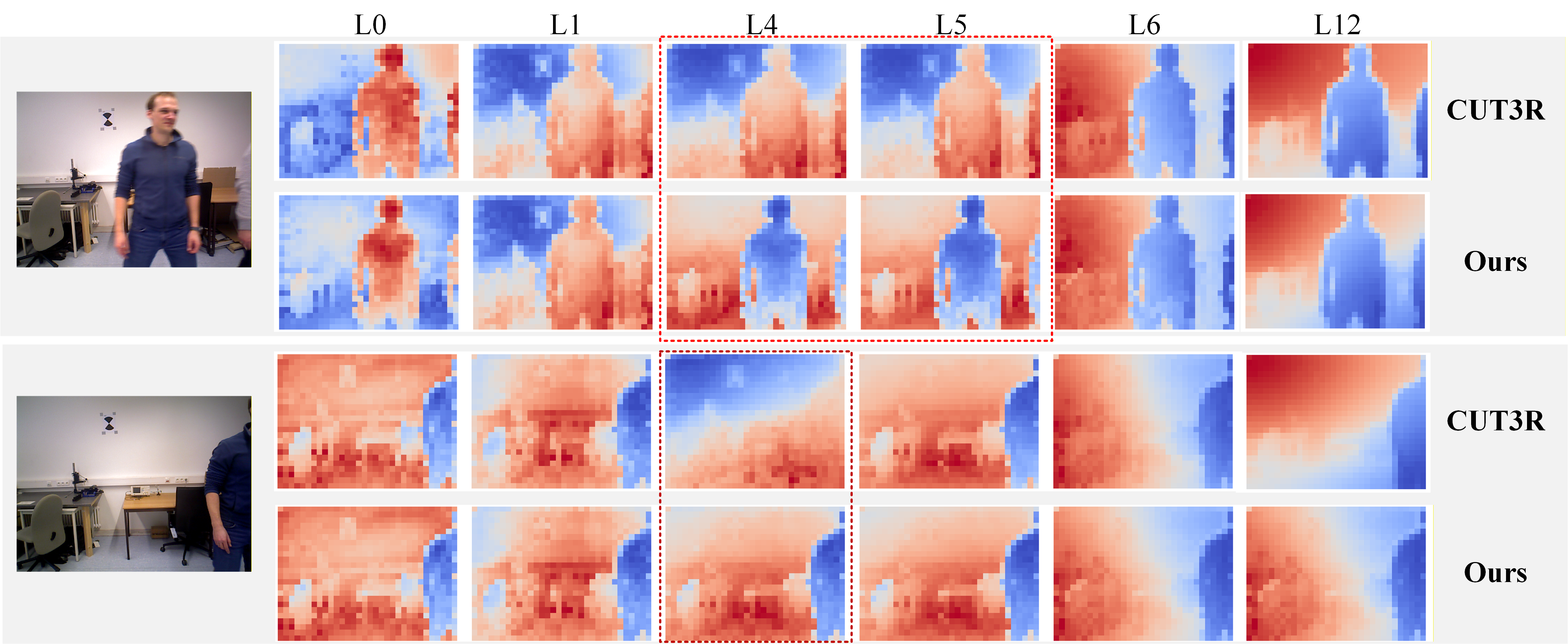}
    \caption{
        \textbf{Layer-wise PCA visualization of decoder embeddings before and after early-layer suppression.}
        We visualize the evolution of image-token embeddings across different decoder layers, 
        where $\text{L}k$ represents the embeddings after passing through the $k$-th decoder layer. 
    }
    \label{pca}
    \vspace{-0.5cm}
\end{figure*}

\noindent\textbf{Layer-wise evolution of image token embeddings.}
Beyond the gating formulation, it is crucial to understand how dynamic interference
propagates through the decoder and how early-layer suppression alters this behavior. For both CUT3R and our method, we extract high-dimensional image features $(f_t \in \mathbb{R}^{H \times W \times C})$  at multiple decoder layers
and project them onto the first three PCA components, which are then averaged to form 
a single-channel intensity map that is visualized using a pseudo-color scheme,
as shown in Figure~\ref{pca}.
This visualization reflects the magnitude and spatial coherence of image token embeddings across layers.
Shallow layers produce highly fragmented patterns with irregular activations, reflecting 
localized and unstable feature responses.  
As we move to mid layers, these patterns become smoother and more spatially organized, 
indicating that the decoder begins to integrate broader contextual information.  
In deeper layers, the intensity maps appear globally more coherent, 
although small localized variations remain visible around dynamic regions.

This progressive transition suggests that early decoder layers rely more heavily 
on appearance cues and exhibit stronger sensitivity to scene dynamics, 
whereas deeper layers produce more geometry-consistent and stable representations.  
Therefore, applying our motion-aware gating to the first few attention layers 
helps suppress unstable responses before they propagate through the network, 
allowing deeper layers to form more consistent geometric embeddings.

As shown in Figure~\ref{pca}, in the shallow decoder layers, our method shifts the feature responses
toward stable static regions more quickly, leading to
faster and more coherent feature aggregation.  
Across the entire depth of the decoder, our embeddings evolve more smoothly, while CUT3R occasionally exhibits abrupt changes 
in activation patterns.  
This observation supports our initial hypothesis that early-layer attention inherently captures motion instability,
and suppressing it leads to more stable geometric reasoning without retraining.


%% file: sec/4_experiment.tex
\section{Experiment}


We evaluate our method on a variety of tasks, including video depth evaluation (Section~\ref{sec:video_depth_evaluation}), camera pose evaluation (Section~\ref{sec:camera_pose_evaluation}) and 4D reconstruction (Section~\ref{sec:4d_reconstruction_evaluation}).

\paragraph{Baselines.} 
We compare our approach with a broad range of transformer-based 3D reconstruction methods.
For global or pairwise-view baselines, we include DUSt3R~\cite{wang2024dust3r}, MASt3R~\cite{leroy2024grounding}, MonST3R~\cite{zhang2025monst3r}, and Easi3R~\cite{chen2025easi3r}, which perform offline inference by jointly optimizing over image pairs.
We also consider Fast3R~\cite{yang2025fast3r} and VGG-T~\cite{wang2025vggt}, which rely on bi-directional attention and therefore operate in a non-streaming, global manner.
For streaming-based comparison, we evaluate against Spann3R~\cite{wang20253d}, CUT3R~\cite{wang2025continuous}, Point3R~\cite{wu2025point3r}, and STREAM3R~\cite{lan2025stream3r}, all of which process video causally by maintaining frame-to-frame continuity through
either recurrent states or cached feature memories. STREAM3R provides two variants: $\alpha$ (DUSt3R-based) and $\beta$ (VGG-T-based), where only the $\beta$ model is currently publicly released.
In our reporting, global and streaming methods are grouped separately to contrast offline optimization with online, frame-by-frame inference. In the table, we categorize methods by Type, indicating whether they are optimization-based (Optim), streaming (Stream), or full-attention (FA).


\begin{table*}[t]
\centering
\small
\setlength{\tabcolsep}{4pt}
\begin{tabular}{l l l | cc | cc | cc}
\toprule
\multirow{2}{*}{\textbf{Alignment}} &
\multirow{2}{*}{\textbf{Method}} &
\multirow{2}{*}{\textbf{Type}} &
\multicolumn{2}{c|}{\textbf{Sintel}} &
\multicolumn{2}{c|}{\textbf{Bonn}} &
\multicolumn{2}{c}{\textbf{KITTI}} \\
& & &
AbsRel$\downarrow$ & $\delta<1.25\uparrow$ &
AbsRel$\downarrow$ & $\delta<1.25\uparrow$ &
AbsRel$\downarrow$ & $\delta<1.25\uparrow$ \\
\midrule

\multirow{10}{*}{\textbf{Per-sequence scale}}
& VGG-T~\cite{wang2025vggt} & FA & \textbf{0.297} & \textbf{68.8} & \textbf{0.055} & \textbf{97.1} & \textbf{0.073} & \textbf{96.5} \\
& Fast3R~\cite{yang2025fast3r} & FA & 0.653 & 44.9 & 0.193 & 77.5 & \underline{0.140} & \underline{83.4} \\
& DUSt3R-GA~\cite{wang2024dust3r} & Optim & 0.656 & 45.2 & 0.155 & 83.3 & 0.144 & 81.3 \\
& MASt3R-GA~\cite{leroy2024grounding} & Optim & 0.641 & 43.9 & 0.252 & 70.1 & 0.183 & 74.5 \\
& MonST3R-GA~\cite{zhang2025monst3r} & Optim & \underline{0.378} & \underline{55.8} &\underline{0.067} & \underline{96.3} & 0.168 & 74.4 \\

\cmidrule(lr){2-9}
& Spann3R~\cite{wang20253d} & Stream &
0.622 & 42.6 &
0.144 & 81.3 &
0.198 & 73.7 \\

& STREAM3R$^{\alpha}$~\cite{lan2025stream3r} & Stream &
0.478 & \textbf{51.1} &
0.075 & 94.1 &
\textbf{0.116} & \textbf{89.6} \\

& Point3R~\cite{wu2025point3r} & Stream &
0.452 & \underline{48.9} &
\textbf{0.060} & \textbf{96.0} &
0.136 & 84.2 \\

& CUT3R~\cite{wang2025continuous} & Stream &
\textbf{0.421} & 47.9 &
0.078 & 93.7 &
\underline{0.118} & 88.1 \\

& \textbf{Ours} & Stream &
\underline{0.451} & 48.6 &
\underline{0.070} & \textbf{96.2} &
\textbf{0.116} & \underline{88.3} \\
\midrule

\multirow{5}{*}{\textbf{Metric scale}}
& MASt3R-GA~\cite{leroy2024grounding} & Optim & 1.022 & 14.3 & 0.272 & 70.6 & 0.467 & 15.2 \\
& STREAM3R$^{\alpha}$~\cite{wang20253d} & Stream & 1.041 & 21.0 & \textbf{0.084} & 94.4 & 0.234 & 57.6 \\
& Point3R~\cite{wu2025point3r} & Stream & \textbf{0.777} & 17.1 & 0.137 & \underline{94.7} & 0.191 & 73.8 \\
& CUT3R~\cite{wang2025continuous} & Stream & 1.029 & \underline{23.8} & 0.103 & 88.5 & \textbf{0.122} & \underline{85.5} \\
& \textbf{Ours} & Stream & \underline{0.820} & \textbf{25.2} & \underline{0.086} & \textbf{96.0} & \underline{0.125} & \textbf{85.8} \\
\bottomrule
\end{tabular}

\caption{\textbf{Video Depth Evaluation} on the
Sintel~\cite{butler2012naturalistic}, Bonn~\cite{palazzolo2019refusion}, and KITTI~\cite{geiger2012we} datasets. We evaluate scale-invariant and metric depth accuracy. Lower AbsRel and higher $\delta<1.25$ indicate better results.
The best and second best results are shown in \textbf{bold} and \underline{underlined}.
}
\label{tab:videodepth}
\end{table*}

\begin{table*}[t]
\centering
\small
\setlength{\tabcolsep}{4pt}
\begin{tabular}{l c | ccc | ccc | ccc}
\toprule
\multirow{2}{*}{\textbf{Method}} &
\multirow{2}{*}{\textbf{Type}} &
\multicolumn{3}{c|}{\textbf{Sintel}} &
\multicolumn{3}{c|}{\textbf{TUM-dynamics}} &
\multicolumn{3}{c}{\textbf{ADT}} \\
& & 
ATE$\downarrow$ & RPE$_{\text{trans}}$$\downarrow$ & RPE$_{\text{rot}}$$\downarrow$ &
ATE$\downarrow$ & RPE$_{\text{trans}}$$\downarrow$ & RPE$_{\text{rot}}$$\downarrow$ &
ATE$\downarrow$ & RPE$_{\text{trans}}$$\downarrow$ & RPE$_{\text{rot}}$$\downarrow$ \\
\midrule
DUSt3R~\cite{wang2024dust3r} & Optim &
0.417 & 0.250 & 5.796 &
\underline{0.083} & \underline{0.017} & 3.567 &
\underline{0.042} & \underline{0.025} & 1.212 \\

MASt3R~\cite{leroy2024grounding} & Optim &
0.185 & 0.060 & 1.496 &
\textbf{0.038} & \textbf{0.012} & \textbf{0.448} &
\textbf{0.015} & 0.125 & \textbf{0.354} \\

MonST3R~\cite{zhang2025monst3r} & Optim &
\underline{0.111} & \underline{0.044} & \underline{0.869} &
0.098 & 0.019 & \underline{0.935} &
0.055 & \underline{0.025} & 0.776 \\

Easi3R~\cite{chen2025easi3r} & Optim &
\textbf{0.110} & \textbf{0.042} & \textbf{0.758} &
0.105 & 0.022 & 1.064 &
\underline{0.042} & \textbf{0.015} & \underline{0.655} \\
\midrule

Spann3R~\cite{wang20253d} & Stream &
0.329 & 0.110 & 4.471 &
0.056 & \underline{0.021} & 0.591 &
0.146 & 0.056 & 1.351 \\

Point3R~\cite{wu2025point3r} & Stream &
0.351 & 0.128 & 1.822 &
0.075 & 0.029 & 0.642 &
- & - & - \\
CUT3R~\cite{wang2025continuous} & Stream &
\textbf{0.213} & \underline{0.066} & \textbf{0.621} &
\underline{0.046} & \textbf{0.015} & \underline{0.473} &
\underline{0.084} & \underline{0.025} & \underline{0.490} \\

\textbf{Ours} & Stream &
\underline{0.228} & \textbf{0.062} & \underline{0.751} &
\textbf{0.042} & \textbf{0.015} & \textbf{0.445} &
\textbf{0.058} & \textbf{0.023} & \textbf{0.477} \\
\bottomrule
\end{tabular}

\caption{
\textbf{Camera pose evaluation} on the Sintel~\cite{butler2012naturalistic}, TUM-dynamics~\cite{sturm2012benchmark}, and ADT~\cite{koppula2024tapvid,lowe2004distinctive} dataset.
All metrics are lower-the-better. Best and second-best are \textbf{bold} and \underline{underlined}.
}
 \vspace{-0.3cm}
\label{tab:pose_eval}
\end{table*}


\subsection{Video Depth Evaluation}
\label{sec:video_depth_evaluation}
Following previous methods~\cite{wang2025continuous,zhang2025monst3r,lan2025stream3r}, we evaluate video depth on Sintel~\cite{butler2012naturalistic}, Bonn~\cite{palazzolo2019refusion}, KITTI~\cite{geiger2012we} dataset, which cover indoor and outdoor, realistic and synthetic dynamic data.
We report the standard metrics, including absolute relative error (Abs Rel) and the percentage of inlier pixels within a 1.25 ratio to ground-truth depth ($\delta<1.25$), which together reflect both per-frame accuracy and the temporal consistency of depth across video sequences.
All methods are first evaluated under the per-sequence scale setting.
For metric-scale methods, such as CUT3R and our MUT3R, 
we further report results in the metric scale without alignment to ground truth. 
As shown in Table\ref{tab:videodepth}, across most datasets, our method consistently improves over existing streaming approaches, particularly our baseline CUT3R.
By keeping the decoder focused on stable static structure rather than reacting to transient motion in early layers, our approach reduces noisy state updates and avoids the abrupt feature shifts that often lead to large depth errors.
As a result, the predicted depth becomes more temporally consistent and less sensitive to transient motion across frames.

\subsection{Camera Pose Evaluation}
\label{sec:camera_pose_evaluation}
We evaluate our camera pose estimation on three dynamic-scene benchmarks in Table~\ref{tab:pose_eval}, including Sintel~\cite{butler2012naturalistic}, TUM-dynamics~\cite{sturm2012benchmark}, and ADT~\cite{koppula2024tapvid,lowe2004distinctive}.
Sintel and TUM-dynamics contain significant non-rigid motion that challenges
correspondence-based pose estimation, while ADT consists of egocentric videos
with rapid motion and large viewpoint changes.
We report three standard metrics:
Absolute Translation Error (ATE), Relative Translation Error (RPE$_{\text{trans}}$),
and Relative Rotation Error (RPE$_{\text{rot}}$)~\cite{sturm2012benchmark}.
All metrics are computed after aligning the predicted trajectory to the
ground-truth camera path using a Sim(3) transformation, following the
protocol of~\cite{teed2021droid,zhang2025monst3r,wang2025continuous}. Global methods degrade substantially under motion, and we further observe that in streaming models such as CUT3R, dynamic regions can introduce early-layer fluctuations that affect both the image features and the recurrent state involved in pose estimation.
By mitigating motion-driven noises, we limit their effect on pose-related image tokens as well as on the recurrent state, leading to cleaner cues and more stable trajectories.

\begin{figure*}[t]
    \centering
    \includegraphics[width=\linewidth]{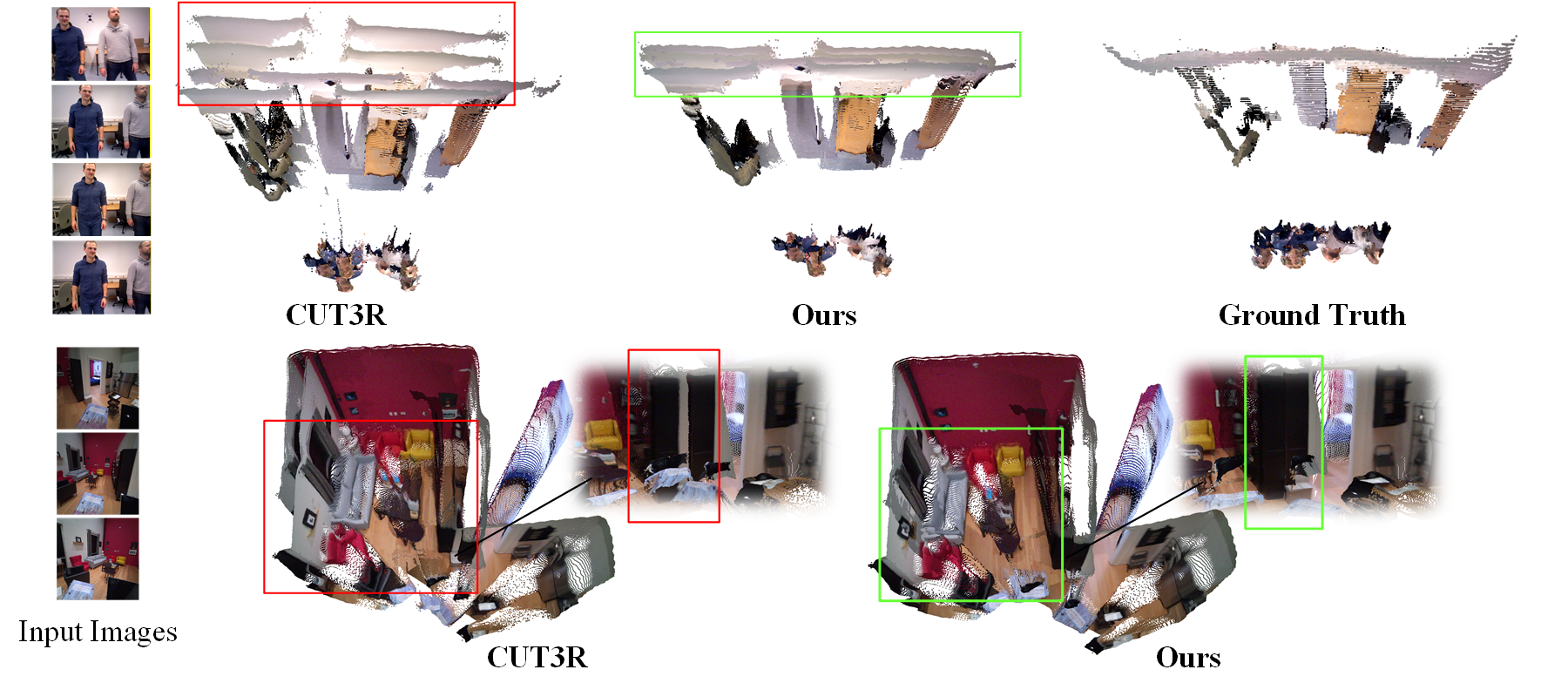}
    \vspace{-0.4cm}
\caption{
\textbf{Qualitative 4D reconstruction comparison} on Bonn and ADT dataset. The region outlined in the image is marked in red to signify lower predictive accuracy, in green to signify higher accuracy.}
\label{fig:reconstruction}
\vspace{-0.4cm}
\end{figure*}

\subsection{4D Reconstruction Evaluation.}
\label{sec:4d_reconstruction_evaluation}
We evaluate 4D reconstruction on the DyCheck~\cite{gao2022monocular} dataset by measuring the geometric distance between the reconstructed and ground-truth point clouds.
DyCheck contains diverse, in-the-wild dynamic videos captured with handheld cameras, making it a challenging benchmark for temporally consistent reconstruction.
Following prior work, we report three standard metrics: \emph{accuracy}, \emph{completeness}, and \emph{overall distance}.
Accuracy measures the nearest-point Euclidean distance from the reconstructed point cloud to the ground truth, 
completeness measures the reverse direction, 
and distance denotes their average, computed via symmetric point matching.
Across all metrics, our method shows clear improvements over CUT3R on DyCheck. By reducing motion-induced feature fluctuations in the early layers, our method produces more stable per-frame point clouds, which in turn leads to smoother and more geometrically consistent surfaces across time.
A qualitative comparison on Bonn and ADT dataset (Figure.~\ref{fig:reconstruction}) further shows that, in dynamic regions where CUT3R tends to drift or lose consistency, our method produces noticeably more stable and temporally coherent geometry.


\begin{table}[t]
\centering
\small
\setlength{\tabcolsep}{3pt}
\begin{tabular}{l | cc | cc | cc}
\toprule
\multirow{2}{*}{\textbf{Method}} &
\multicolumn{2}{c|}{\textbf{Accuracy}$\downarrow$} &
\multicolumn{2}{c|}{\textbf{Completion}$\downarrow$} &
\multicolumn{2}{c}{\textbf{Distance}$\downarrow$} \\
& Mean & Med. & Mean & Med. & Mean & Med. \\
\midrule

\rowcolor{gray!15}
\multicolumn{7}{l}{\em Optimization-based methods} \\

DUS3R~\cite{wang2024dust3r}       & 0.802 & 0.595 & 1.950 & 0.815 & 0.353 & 0.233 \\
MonST3R~\cite{zhang2025monst3r}   & 0.851 & 0.689 & 1.734 & 0.958 & 0.353 & 0.254 \\
DAS3R~\cite{xu2024das3r}          & 1.772 & 1.438 & 2.503 & 1.548 & 0.475 & 0.352 \\
Easi3R~\cite{chen2025easi3r}      & \textbf{0.703} & \textbf{0.589} &
                                   \textbf{1.474} & \textbf{0.586} &
                                   \textbf{0.301} & \textbf{0.186} \\
\midrule

\rowcolor{gray!15}
\multicolumn{7}{l}{\em Streaming methods} \\

CUT3R~\cite{wang2025continuous}   & 0.461 & 0.338 & 1.648 & 0.813 & 0.329 & 0.231 \\
\textbf{Ours}                     & \textbf{0.438} & \textbf{0.331} &
                                   \textbf{1.487} & \textbf{0.663} &
                                   \textbf{0.322} & \textbf{0.189} \\
\bottomrule
\end{tabular}
\vspace{-0.2cm}
\caption{
\textbf{Quantitative comparisons for point cloud reconstruction} on the DyCheck dataset~\cite{gao2022monocular}.
Best results are shown in \textbf{bold}.
}
\vspace{-0.5cm}
\label{tab:dycheck}

\end{table}

%% file: sec/5_ablation_study.tex
\section{Ablation study}

Table~\ref{tab:ablation} reports ablations on the Bonn dataset under the metric-scale 
video depth evaluation protocol.
We first vary the suppression depth of decoder layers and observe that applying suppression to the early-to-middle layers yields better performance, while deeper suppression leads to degraded accuracy. This observation indicates that early-layer attention responses are particularly sensitive to scene dynamics, 
and stabilizing them leads to clear improvements in overall reconstruction quality.
We then evaluate the contribution of each gating configuration. Removing any of the three gating modules--image self-attention, image-to-state, or state-to-image—consistently reduces performance, showing that dynamic noise arises not only from image–state exchanges but, most prominently, from image self-attention, which requires explicit filtering. Combining all gating modules within the optimal depth range yields the best overall performance.

\begin{table}[t]
\centering
\small
\setlength{\tabcolsep}{3pt}
\begin{tabular}{llcc}
\toprule
\textbf{Method} & \textbf{Setting} & Abs Rel$\downarrow$ & $\delta{<}1.25$ $\uparrow$ \\
\midrule

\multirow{1}{*}{CUT3R~\cite{wang2025continuous}}
  & 0, w/o gates      & 0.103  & 88.5 \\
  \midrule
\multirow{4}{*}{Suppression depth}
  & 0-2 layers  & 0.099 & 91.2 \\
  & 0-8 layers  & 0.114 & 89.4 \\
  & 0-12 layers & 0.106 & 91.4 \\
  & 6-12 layers & 0.113 & 86.6 \\
\midrule
\multirow{3}{*}{Gating configuration}
  & w/o img self & 0.111 & 87.2 \\
  & w/o img-to-state  & 0.089 & 94.1 \\
  & w/o state-to-img & 0.093 & 93.7 \\
\midrule
\multirow{1}{*}{Our Full model}
  & 0-6 layers, all gates     & \textbf{0.086} & \textbf{96.0} \\
\bottomrule
\end{tabular}
\vspace{-0.2cm}
\caption{\textbf{Ablation Study.} We evaluate the impact of applying suppression at different depths with all gating types activated, and additionally study the contribution of each gating component by removing them one at a time at a fixed depth of $L=6$.
}
\vspace{-0.2cm}
\label{tab:ablation}
\end{table}

%% file: sec/6_conclusion.tex
\section{Conclusion}
We introduced MUT3R, a training-free method that leverages the motion-sensitive attention patterns already present in pretrained 3D transformers.
By converting these implicit cues into early soft attention gating, MUT3R suppresses motion interference while preserving CUT3R’s geometric reasoning.
Across multiple dynamic-scene benchmarks, it consistently improves depth stability, pose accuracy, and temporal coherence—all without modifying architecture or retraining.  This work provides new insights for the dynamic 3D reconstruction community, highlighting the potential of using the attention-derived motion cue in pretrained 3D transformers.